\documentclass[journal]{IEEEtran}
\ifCLASSINFOpdf
\else
\fi
\usepackage{mathtools}
\usepackage{color}
\usepackage{algorithm}
\usepackage[noend]{algpseudocode}

\usepackage{times}
\usepackage{epsfig}
\usepackage{graphicx}
\usepackage{amssymb}
\usepackage{romannum}
\usepackage{color}

\usepackage{amsmath}
\usepackage{breqn}

\begin{document}
%
\title{Two-path Deep Semi-supervised Learning for Timely Fake News Detection}
%
%
%

 \author{Xishuang Dong,~Uboho Victor, and~Lijun Qian,~\IEEEmembership{Senior Member,~IEEE,}
\thanks{X. Dong, U. Victor, and L. Qian are with the Center of Excellence in Research and Education for Big Military Data Intelligence (CREDIT Center), Department of Electrical and Computer Engineering, Prairie View A\&M University, Texas A\&M University System, Prairie View, TX 77446, USA. Email: xidong@pvamu.edu, uboho.dpc@outlook.com, liqian@pvamu.edu}
}

\maketitle

\begin{abstract}
News in social media such as Twitter has been generated in high volume and speed. However, very few of them are labeled (as fake or true news) by professionals in near real time.
In order to achieve timely detection of fake news in social media,  a novel framework of two-path deep semi-supervised learning is proposed where one path is for supervised learning and the other is for unsupervised learning. The supervised learning path learns on the limited amount of labeled data while the unsupervised learning path is able to learn on a huge amount of unlabeled data. Furthermore, these two paths implemented with convolutional neural networks (CNN) are  jointly optimized to complete semi-supervised learning.  In addition, we build a shared CNN to extract the low level features on both labeled data and unlabeled data to feed them into these two paths. To verify this framework, we implement a Word CNN based semi-supervised learning model and test it on two datasets, namely, LIAR and PHEME. Experimental results demonstrate that the model built on the proposed framework can recognize fake news  effectively with very few labeled data.
\end{abstract}

\begin{IEEEkeywords}
Fake News Detection, Deep Semi-supervised Learning, Convolutional Neural Networks, Joint Optimization
\end{IEEEkeywords}

%
\IEEEpeerreviewmaketitle

\section{Introduction}
\label{sec1}


\begin{figure*} [ht]
	\includegraphics[width=\linewidth]{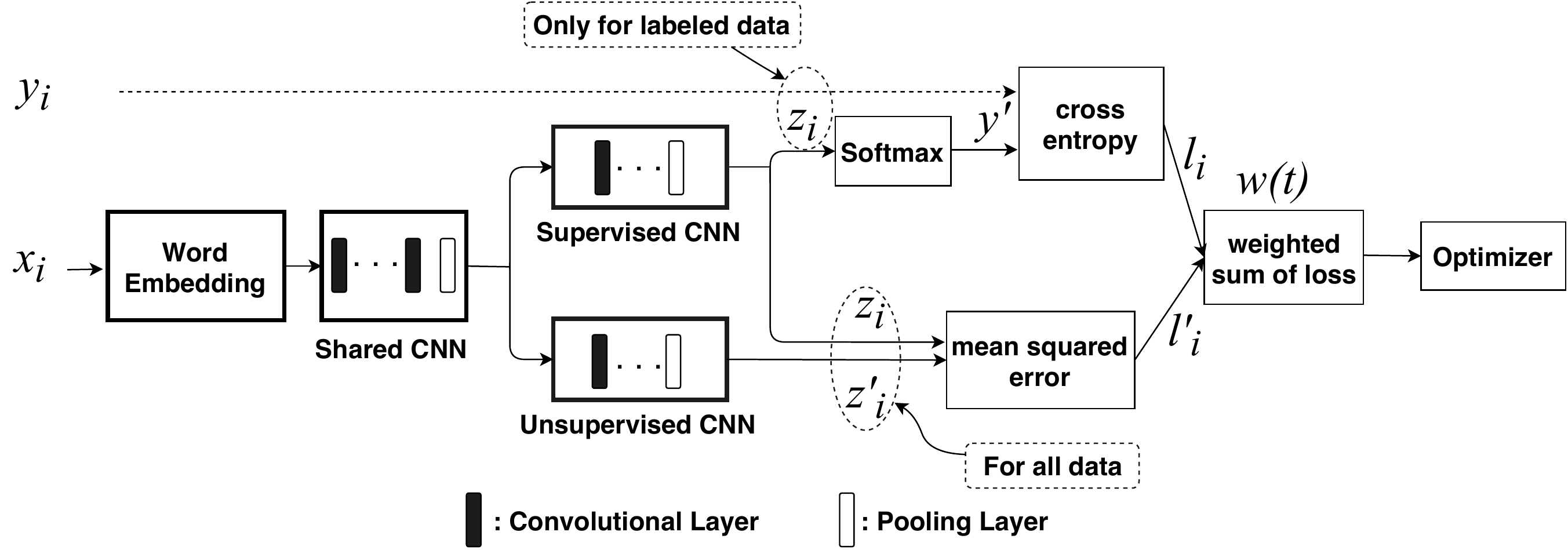}
	\caption{Framework of two-path deep semi-supervised learning. Samples $x_{i}$ are inputs. Labels $y_{i}$ are available only for the labeled inputs and the associated cross-entropy loss component is evaluated only for those. $z_{i}$ and $z'_{i}$ are outputs from the supervised CNN and the unsupervised CNN, respectively. $y'_{i}$ is the predicted label for $x_{i}$. $l_{i}$ is the cross-entropy loss and $l'_{i}$ is the mean squared error loss. $w(t)$ are the weights for joint optimization of $l_{i}$ and $l'_{i}$. 
	}
	\label{Fig1_framework}
\end{figure*}

Social media (e.g., Twitter and Facebook) has become a new ecosystem for spreading news \cite{pennycook2019fighting}. Nowadays, people are relying more on social media services rather than traditional media because of its advantages such as social awareness, global connectivity, and real-time sharing of digital information. Unfortunately, social media is full of fake news. 
Fake news consists of information that is intentionally and verifiably false to mislead readers, which is motivated by chasing personal or organizational profits~\cite{shu2017fake}. For example, fake news has been propagated on Twitter like infectious virus during the 2016 election cycle in the United States \cite{allcott2017social, grinberg2019fake}. Understanding what can be done to discourage fake news is of great importance.

One of the fundamental steps to discourage fake news would be \emph{timely} fake news detection.  Fake news detection \cite{hovy2016enemy, wang2017liar, potthast2018stylometric} is to determine the truthfulness of the news by analyzing the news contents and related information such as propagation patterns. It attracts a lot of attention to resolve this problem from different aspects, where supervised learning based fake news detection dominates this domain. For instance, Ma \textit{et.al} detects fake news with data representations of the contents that are learned on the labeled news \cite{ma2016detecting}. Early detection is also an effective approach to recognize fake news by identifying the signature of text phrases in social media posts \cite{zhao2015enquiring}. Moreover, temporal features play a crucial role in the fast-paced social media environment because information spreads more rapidly than traditional media \cite{kwon2013prominent}. For example, detecting a burst of topics on social media can capture the variations in temporal patterns of news~\cite{chang2016lifecycle, chang2014ups}. Specifically, deep learning based fake news detection achieves the state-of-the-art performance on different datasets \cite{oshikawa2018survey}, where both recurrent neural networks (RNN) and convolutional neural networks (CNN) are employed to recognize fake news \cite{wang2017liar, rashkin2017truth, long2017fake}. However, since news spreads on social media at very high speed when an event happens, only very limited labeled data is available in practice for fake news detection, which is inadequate  for the supervised model to perform well.

As an emerging task in the field of natural language processing (NLP), fake news detection requires big labeled data to meet the requirement of building supervised learning based detection models. However, annotating the news on social media is too expensive and costs a huge amount of human labor due to the huge size of social media data. Furthermore, this is almost impossible to achieve in near real time. In addition, it is difficult to ensure the annotation consistency for big data labeling~\cite{bosco2000building}. With the increment of the data size, the annotation inconsistence will be worse. Therefore, using unlabeled data to enhance fake news detection becomes a promising solution and more urgent.

In this paper, we propose a deep semi-supervised learning framework by building two-path convolutional neural networks to accomplish timely fake news detection in the case of limited labeled data, where the framework is shown in Figure \ref{Fig1_framework}. It consists of three components, namely, a shared CNN, a supervised CNN, and an unsupervised CNN. One path is composed of the shared CNN and supervised CNN while the other is made of the shared CNN and unsupervised CNN. Moreover, the architectures of these three CNNs can be similar or different, which are determined by the application and performance. All data (labeled and unlabeled data) will be used to generate the mean squared error loss, while only labeled data will be used to calculate the cross-entropy loss. Then a weighted sum of these two losses is used to optimize the proposed framework.
We validate the proposed framework on detecting fake news using two datasets, namely, LIAR~\cite{wang2017liar} and PHEME~\cite{zubiaga2016analysing}. Experimental results demonstrate the effectiveness of the proposed framework even with very limited labeled data.

In summary, the contributions of this study are as follows:
\begin{itemize}

\item We propose a novel two-path deep semi-supervised learning (TDSL) framework containing three CNNs, where both labeled data and unlabeled data are used jointly to train the model and enhance the detection performance. 

\item We implement a Word CNN~\cite{kim2014convolutional} based TDSL to detect fake news with limited labeled data and compare its performance with various deep learning based baselines. Specifically, we validate the implemented model by testing on the LIAR  and PHEME datasets.  It is observed that the proposed model detects fake news effectively even with extremely limited labeled data. 

\item The proposed framework could be applied to address other tasks. Furthermore, novel deep semi-supervised learning models can be implemented based on the proposed framework with various designs of CNNs, which will be determined by the intended applications and tasks.

\end{itemize}

\begin{algorithm*}[ht]
	\caption{Learning in the proposed framework}
	\begin{algorithmic}[1]
		\Require{$x_{i}$ = training sample}
		\Require{$S$ = set of training samples}
		\Require{$y_{i}$ = label for labeled $x_{i}$ $i \in S$}
		\Require{$f_{embedding}(x)$ = word embedding}
		\Require{$f_{\theta_{shared}}(x)$ = shared CNN with trainable parameters  $\theta_{shared}$}
		\Require{$f_{\theta_{sup}}(x)$ = supervised CNN with trainable parameters  $\theta_{sup}$}
		\Require{$f_{\theta_{unsup}}(x)$ = unsupervised CNN with trainable parameters  $\theta_{unsup}$}
		\Require{$w(t)$ = unsupervised weight ramp-up function}
		
		\For{$t$ in~[1, num epochs] }
 			 \For{each minibatch $B$}
			 	\State{$x'_{i \in B} \gets f_{embedding}(x_{i \in B})$} \hspace{34mm} $\triangleright$ represent words with word embedding
      				\State{$z_{i \in B} \gets f_{\theta_{sup}}({f_{\theta_{shared}}{(x'_{i \in B})}})$} \hspace{29mm} $\triangleright$ evaluate supervised cnn outputs for  inputs
				\State{$z'_{i \in B} \gets f_{\theta_{unsup}}({f_{\theta_{shared}}{(x'_{i \in B})}})$} \hspace{27mm}$\triangleright$ evaluate unsupervised cnn outputs for  inputs
				\State{$l_{i \in B} \gets -\frac{1}{|B|} \sum_{i \in B \cap S}{log f_{softmax}{(z_{i})}[y_{i}]}$}\hspace{13.5mm}$\triangleright$ supervised loss component
				\State{$l'_{i \in B} \gets \frac{1}{C|B|} \sum_{i \in B}{||z_{i} - z'_{i}||^{2}}$} \hspace{28mm} $\triangleright$ unsupervised loss component
				\State{$loss \gets l_{i \in B} + w(t) \times l'_{i \in B}$} \hspace{33mm} $\triangleright$ total loss 
				\State{update $\theta_{shared}$,  $\theta_{sup}$, $\theta_{unsup}$ using, e.g., ADAM} \quad $\triangleright$ update parameters
 			 \EndFor
		\EndFor
	\Return{$\theta_{shared}$,  $\theta_{sup}$, $\theta_{unsup}$}
	\end{algorithmic}
	\label{Arg1_learning}
\end{algorithm*}

\section{Related Work}
\label{sec3}

Fake news detection has attracted a lot of attention in recent years. There are extensive studies such as content based method and propagation pattern based method. Content based method typically involves two steps: preprocessing news contents and training supervised learning model on the preprocessed contents. The first step usually involves tokenization, stemming, and/or weighting words \cite{manning1999foundations, tong2001support}. In the second step, Term Frequency-Inverse Document Frequency (TF-IDF)~\cite{ramos2003using, paik2013novel} may be employed to build samples to train supervised learning models. 
However, the samples generated by TF-IDF will be sparse, especially for social media data. To overcome this challenge, word embedding methods such as word2vec \cite{mikolov2013efficient} and GloVe \cite{pennington2014glove} are used to convert words into vectors. 

In addition, Mihalcea \textit{et al.} \cite{mihalcea2009lie} used linguistic inquiry and word count (LIWC) \cite{pennebaker2001linguistic} to explore the difference of word usage between deceptive language and non-deceptive ones. Specifically, deep learning based models have been explored more than other supervised learning models~\cite{oshikawa2018survey}. For example, Rashkin \textit{et al.} \cite{ruchansky2017csi} built the detection model with two LSTM RNN models: one learns on simple word embeddings, and the other enhances the performance by concatenating long short-term memory (LSTM) outputs with LIWC feature vectors. Doc2vec \cite{le2014distributed} is also applied to represent content that is related to each social engagement. Attention based RNNs are employed to achieve better performance as well. Long \textit{et al.} \cite{long2017fake} incorporates the speaker names and the statement topics into the inputs to the attention based RNN. In addition, convolutional neural networks are also widely used since they succeed in many text classification tasks. Karimi \textit{et al.} \cite{karimi2018multi} proposed Multi-source Multi-class Fake news Detection framework (MMFD), where CNN analyzes local patterns of each text in a claim and LSTM analyze temporal dependencies in the entire text. 

In propagation pattern based method, the propagation patterns have been extracted from time-series information of news spreading on social media such as Twitter and they are used as features for detecting fake news. For instance, to identify fake news from microblogs, Ma \textit{et al.} \cite{ma2015detect} proposed the Dynamic Series-Time Structure (DSTS) to capture variations in social context features such as microblog contents and users over time for early detection of rumors.  Lan \textit{et al.}  \cite{lan2018mining} proposed Hierarchical Attention RNN (HARNN) that uses a Bi-GRU layer with the attention mechanism to capture high-level representations of rumor contents, and a gated recurrent unit (GRU) layer to extract semantic changes.  Hashimoto \textit{et al.}  \cite{hashimoto2011rumor}  visualized topic structures in time-series variation and seeks help from external reliable source to determine the topic truthfulness. 

In summary, most of the current methods are based on supervised learning. It requires a large amount of labeled data to implement the detection processes, especially for the deep learning based approaches. However, annotating the news on social media is too expensive and costs a huge amount of human labor due to the huge size of social media data. Furthermore, this is almost impossible to achieve in near real time. Even with labeled data, constructing the huge amount of labeled corpus is an extremely difficult task in the field of natural language processing as it costs a large volume of resources and it is challenging to guarantee the label consistence. Therefore, it is imperative to incorporate  unlabeled data together with labeled data in fake news detection to enhance the detection performance. Semi-supervised learning \cite{chapelle2009semi, zhu2005semi} is a technique that is able to use both labeled data and unlabeled data. Therefore, we propose a novel framework of deep semi-supervised learning via convolutional neural networks for timely fake news detection. In the next section, we present the proposed framework and show how to implement the deep semi-supervised learning model in detail.

\section{Model}
\label{sec4}
\begin{figure*} [ht]
	\includegraphics[width=\linewidth]{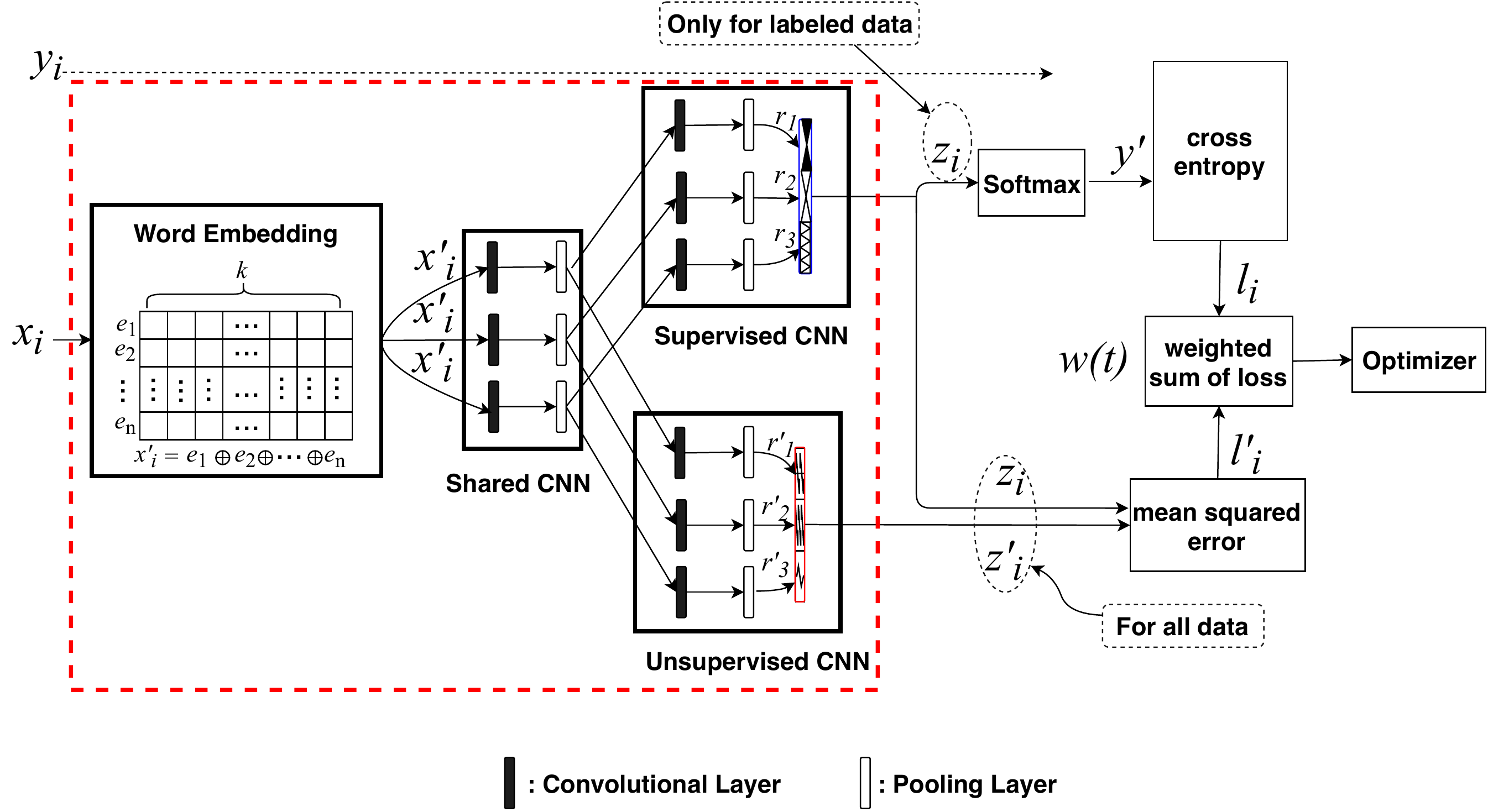}
	\caption{Word-CNN Based Deep Semi-supervised Learning. In the shared CNN, each convolutional layer contains 100 ($3\times3$) filters, 100 ($4\times4$) filters, and 100 ($5\times5$) filters, respectively. Both the supervised CNN and the unsupervised CNN have the same architecture of the shared CNN with different numbers of filters, where each convolutional layer contains 100 ($3\times3$) filters.  We use ($2\times2$) max-pooling for all pooling layers.  $\oplus$ is the concatenation operator.  $r_{1}$, $r_{2}$ and $r_{3}$ are outputs from the supervised path while $r'_{1}$, $r'_{2}$ and $r'_{3}$ are those from the unsupervised path. Furthermore, we concatenate $r_{1}$, $r_{2}$ and $r_{3}$ to conduct $z_{i}$ and connect $r'_{1}$, $r'_{2}$ and $r'_{3}$ to generate $z'_{i}$.}
	\label{Fig_implementation}
\end{figure*}

We propose a general framework of deep semi-supervised learning and apply it to accomplish fake news detection. Suppose the training data consist of total $N$ inputs, out of which $M$ are labeled. The inputs, denoted by $x_{j}$, where $ j \in {1... N}$, are the news contents that contain sentences related to fake news. In general, the news on social media normally contains limited number of words like 100 or less. It will lead to the data sparsity if we apply TF-IDF to extract features. To relieve the data sparsity problem, we employ word embedding techniques, for instance, word2vec \cite{mikolov2013efficient, mikolov2013distributed, goldberg2014word2vec},  to represent the news contents. Here $S$ is the set of labeled inputs, $|S| = M$. For every $i \in S$, we have a known correct label $y_{i} \in {1... C}$, where $C$ is the number of different classes. 

The proposed framework of deep semi-supervised learning and corresponding learning procedures are shown in Figure \ref{Fig1_framework} and Algorithm \ref{Arg1_learning}, respectively.  As shown in Figure \ref{Fig1_framework}, we evaluate the network for each training input $x_{i}$ with the supervised path and the unsupervised path to complete two tasks, resulting in prediction vectors $z_{i}$ and $z'_{i}$, respectively. One task is to learn how to mine patterns of fake news regarding the news labels while the other is to optimize the representations of news without the news labels.   Specially, before these two paths, there is a shared CNN to extract low-level features to feed the later two CNNs. It is similar to deep multi-task learning \cite{zhang2014facial, ruder2017overview}  since the low-level features are shared in the different tasks \cite{chowdhury2018multitask, dong2019deep}.  The major difference between the proposed framework and deep multi-task learning is that tasks in the proposed framework will involve both supervised learning and unsupervised learning, while all tasks in deep multi-task learning are only based on supervised learning. 

In addition, these two paths can have independent CNNs with the identical or different setups for supervised learning and unsupervised learning, respectively. They generate two prediction vectors that are new representations for the inputs with respect to their tasks. For the identical setups of these two path, i.e., using the same CNN structure for both paths, it is important to notice that, because of dropout regularization, training CNNs in these two paths is a stochastic process. This will result in the two CNNs having different link weights and  filters during training. It implies that there will be difference between the two prediction vectors $z_{i}$ and $z'_{i}$ of the same input $x_{i}$. Given that the original input $x_{i}$ is the same, this difference can be seen as an error and thus minimizing the mean square error (MSE) is a reasonable objective in the learning procedure. 

We utilize those two vectors $z_{i}$ and $z'_{i}$ to calculate the loss given by 
\begin{dmath}
Loss = -\frac{1}{|B|} \sum_{i \in B \cap S}{log f_{softmax}{(z_{i})}[y_{i}]} 
+ w(t) \times \frac{1}{C|B|} \sum_{i \in B}{||z_{i} - z'_{i}||^{2}} \; ,
\label{Equ1_loss}
\end{dmath}
where $B$ is the minibatch in the learning process. The loss consists of two components. As illustrated in Algorithm \ref{Arg1_learning}, $l_{i}$ is the standard cross-entropy loss to evaluate the loss for labeled inputs only. On the other hand, $l'_{i}$, evaluated for all inputs, penalizes different predictions for the same training input $x_{i}$ by taking the mean squared error between $z_{i}$ and $z'_{i}$. To combine the supervised loss $l_{i}$ and unsupervised loss $l'_{i}$, we scale the latter by time-dependent weighting function $w(t)$ \cite{laine2016temporal} that ramps up, starting from zero, along a Gaussian curve. In the beginning of training, the total loss and the learning gradients are dominated by the supervised loss component, i.e., the labeled data only. Unlabeled data will contribute more than the labeled data at later stage of training.

Based on the proposed framework, we implement a deep semi-supervised learning model with Word CNN \cite{kim2014convolutional}, where the architecture is shown in Figure \ref{Fig_implementation}. The Word CNN is a powerful classifier with the simple architecture of CNN for sentence classification \cite{kim2014convolutional}. Considering fake news detection, let $e_{i} \in \mathbb{R}^{k}$ be the $k$-dimensional word vector corresponding to the $i$-th word of the sentence in the news. A sentence of length $n$ is represented as
\begin{dmath}
x'_{i} = e_{1} \oplus e_{2} \oplus ... \oplus e_{n},
\label{Equ2_sentence}
\end{dmath}
where $ \oplus$ is the concatenation operator. 
A convolution operation involves a filter $c \in \mathbb{R}^{hk}$, which is applied to a window of $h$ words to produce a new feature. The pooling operation deals with variable sentence lengths. As shown in Figure \ref{Fig_implementation}, the input $x_{i}$ is represented as $x'_{i}$ with the word embedding\footnote{https://www.tensorflow.org/api\_docs/python/tf/nn/embedding\_lookup}. Then the representation $x'_{i}$ is input into three convolutional layers followed by pooling layers to extract low-level features in the shared CNN. The extracted low-level features are given to the supervised CNN and unsupervised CNN, respectively. Moreover, the outputs from pooling layers in these two paths are concatenated as two vectors $z_{i}$ and $z'_{i}$, respectively. For the supervised path, we use the label $y_{i} $ and the result generated by running softmax on $z_{i}$ to build the supervised loss with the cross entropy. On the other hand, these two vectors $z_{i}$ and $z'_{i}$ are employed to construct the unsupervised loss with the mean squared error. Finally, the supervised loss and unsupervised loss are jointly optimized with the Adam optimizer, which is shown in Algorithm \ref{Arg1_learning}.

Although there are a few related works in the literature such as the $\Pi$ model \cite{laine2016temporal}, there exist significant differences. Firstly, the $\Pi$ model is designed for processing image data while the proposed framework is more suitable to be used to solve problems related to NLP. Secondly, in the $\Pi$ model, it combined image data augmentation with dropout to generate two outputs, but it cannot be used to process language in NLP tasks since data augmentation for processing text contents is difficult to implement. Finally, instead of using one path CNN, we apply two independent CNNs to generate those two outputs. Furthermore, the proposed model is more flexible as the two independent paths can be tuned for specific goals.

\section{Experiment}
\label{sec5}
\subsection{Datasets}

To validate the effectiveness of the proposed framework on fake news detection, we test the implemented model on two benchmarks: LIAR and PHEME.

\begin{figure*} [ht]
	\includegraphics[scale=0.5]{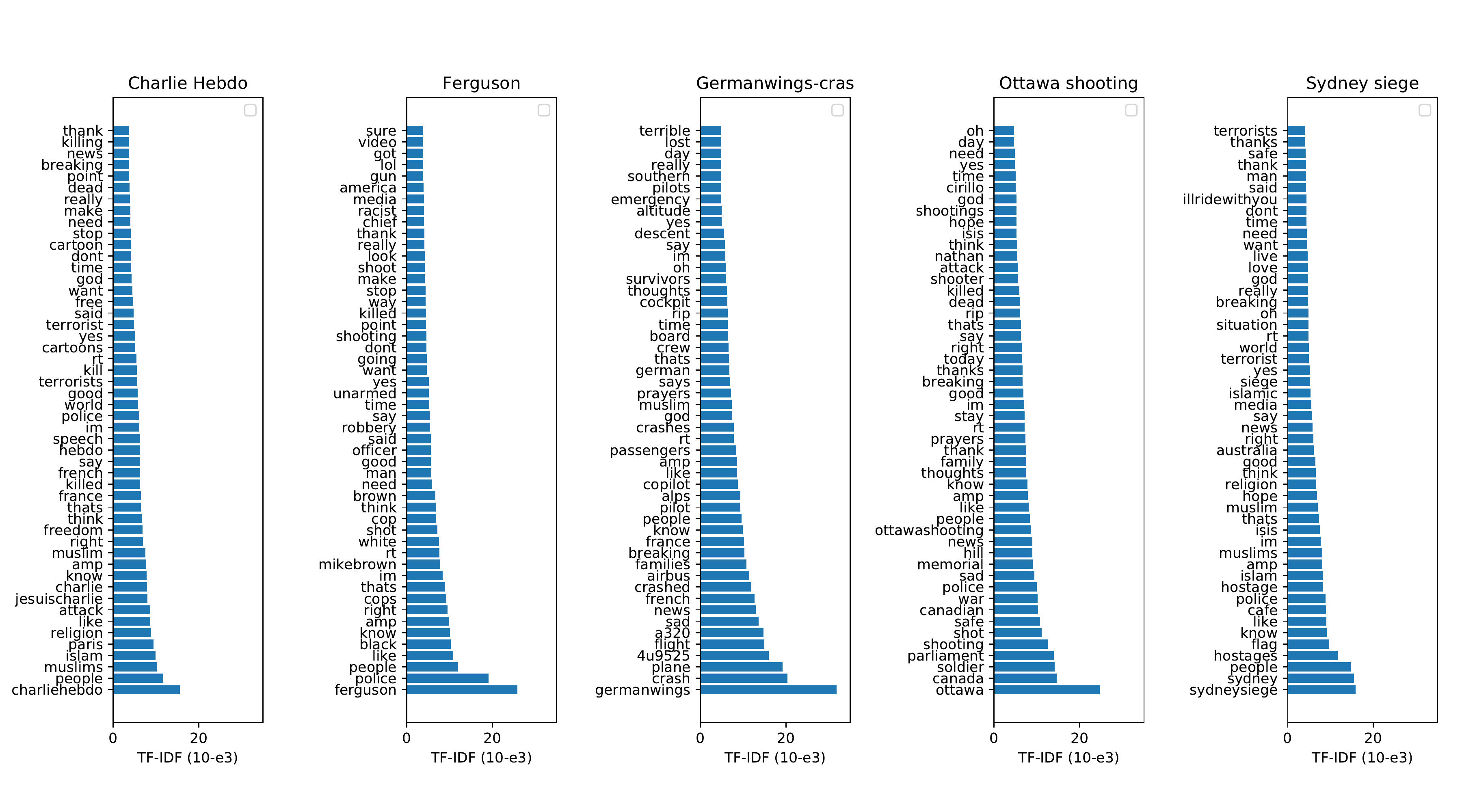}
	\centering
	\caption{An  example of the difference of word distributions between five events in PHEME dataset. x-axis indicates the TF-IDF value of the word while y-axis shows top 50 words ranked by corresponding TF-IDF values.}
	\label{Fig_word_distribution}
\end{figure*}


\subsubsection{LIAR}LIAR \cite{wang2017liar} is the recent benchmark dataset for fake news detection. This dataset includes 12,836 real-world short statements collected from PolitiFact\footnote{https://www.politifact.com}, where editors handpicked the claims from a variety of occasions such as debate, campaign, Facebook, Twitter, interviews, ads, etc. Each statement is labeled with six-grade truthfulness, namely, true, false, half-true, part-fire, barely-true, and mostly-true. The information about the subjects, party, context, and speakers are also included in this dataset. In this paper,  this benchmark contains three parts: training dataset with 10,269 statements, validation dataset with 1,284 statements, and testing dataset with 1,283 statements. Furthermore, we reorganize the data as two classes by treating five classes including false, half-true, part-fire, barely-true, and mostly-true as Fake class and true as True class. Therefore, the fake news detection on this benchmark becomes a binary classification task.

\subsubsection{PHEME}We employ PHEME \cite{zubiaga2016analysing} to verify the effectiveness of the proposed framework on social media data, where it collects 330 twitter threads. Tweets in PHEME are labeled as true or false in terms of thread structures and follow-follower relationships. PHEME dataset is related to nine events whereas this paper only focuses on the five main events, namely, Germanwings-crash (GC), Charlie Hebdo (CH), Sydney siege (SS), Ferguson (FE), and Ottawa shooting (OS). It has different levels of annotations such as thread level and tweet level. We only adopt the annotations on the tweet level and thus classify the tweets as fake or true.  The detailed distribution of tweets and classes is shown in table \ref{tab1_PHAME} after the data is preprocessed such as removing data redundancy. It is observed that the class distribution is different among these events. For example, three events including CH, SS, and FE have more true news while the event GC and OS have more fake news. Furthermore, class distributions for events such as CH, SS, FE are significantly imbalanced, which will be a challenge to the detection task (a binary classification task).

In addition, the word distributions vary among five events, where  one example is shown in Figure \ref{Fig_word_distribution}: there are 50 words ranked by their TF-IDF values, where TF-IDF values are used to evaluate the word relevance to the event \cite{rajaraman2011mining}.  It is observed that the top ranked words are very different for different events. Moreover, even for the same word, their relevance are not the same for different events. For example, the relevance (TF-IDF values) of the word ``\textit{thank}" are different between events including Charlie Hebdo (CH), Ferguson (FE), and Ottawa shooting (OS).

In addition, we perform leave-one-event-out (LOEO) cross-validation \cite{kochkina2018all}, which is closer to the realistic scenario where we have to verify unseen truth. For example, the training data can contain the events GC, CH, SS, and FE whereas the testing data will contain the event OS. However, as highlighted in Figure \ref{Fig_word_distribution}, this fake news detection task becomes very difficult since the training data has very different data distribution from that of the testing data,. 

It should be noted that \emph{the data in original datasets is fully labeled}. We employ \emph{all labeled data to train the baseline models}. Compared to the baselines, \emph{we train the proposed models with only partially labeled data}  and the percentage of labeled data for training of our proposed models is from 1\% to 30\%.

\subsection{Experiment Setup}

The key hyper-parameters for the implemented model are shown in Table \ref{tab_hyperpara} and we employ Adam optimizer to complete the training the implemented proposed model (TDSL).

\begin{table}
	\caption{\label{tab_hyperpara} Hyper-parameters for the training of Word CNN based TDSL}
        \begin{center}
                \begin{tabular}{|l|r|}
                 \hline \textbf{Hyper-parameters} & \textbf{Values} \\ \hline
                	Dropout		& 0.5 \\
		Minibatch size	& 128 \\
		Number of epochs	& 200 \\
		Maximum learning rate & 1e-4 \\
		\hline
               \end{tabular}
       \end{center}
 \end{table}

In addition, we employ five deep supervised learning models as baselines including 1) Word-level CNN (Word CNN) \cite{kim2014convolutional}, 2) Character-level CNN (Char CNN)~\cite{zhang2015character}, 3) Very Deep CNN (VD CNN) \cite{conneau2016very}, 4) Attention-Based Bidirectional RNN (Att RNN)~\cite{zhou2016attention},  and 5) Recurrent CNN (RCNN)~\cite{lai2015recurrent}, where these models perform well on text classification. Specifically, Word CNN performs well on sentence classification, which is more suitable to process social media data as the length of the content of the data is short like that of the sentence. In addition, we build 6) word-level bidirectional RNN (Word RNN) to compare the implemented model, where Word RNN contains one embedding layer and one bidirectional RNN layer, and concatenate all the outputs from the RNN layer to feed to the final layer that is a fully-connected layer. Thus, there are total 6 baseline models. \emph{Note that baseline models use all labeled data from the original datasets}.

\begin{table}
	\caption{\label{tab1_PHAME} Number of tweets and class distribution in the PHEME dataset.}
        \begin{center}
                \begin{tabular}{|l|rrr|}
                    \hline \textbf{Events} & \textbf{Tweets} & \textbf{Fake} & \textbf{True}\\ \hline
                    	   Germanwings-crash	& 3,920 	& 2,220	& 1,700 \\
                            Charlie Hebdo		& 34,236  	& 6,452	& 27,784 \\
                            Sydney siege			& 21.837 	& 7,765	& 14,072 \\
                            Ferguson			& 21,658	& 5,952	& 15,706 \\ 
                            Ottawa shooting		& 10,848	& 5,603	& 5,245 \\ 
                            \hline
                            \textbf{Total} 	& \textbf{92,499}	& \textbf{27,992}	& \textbf{64,507} \\  
                     \hline
                \end{tabular}
       \end{center}
        
\end{table}

\subsection{Evaluation}

We apply different  metrics to evaluate the performance of fake news detection regarding the task features on these two benchmarks. 

\begin{itemize}

\item \textbf{LIAR: }  We employ accuracy, precision, recall and Fscore to evaluate the detection performance. Accuracy is calculated by dividing the number of statements detected correctly over the total number of statements. 

\begin{equation}
	Accuracy = \frac{N_{correct}}{N_{total}}.
\end{equation}

In addition, we employ $Fscore$ values of each class to check the performance since the task is a binary text classification.

\begin{equation}
	Fscore = \frac{2 \times Precision \times Recall}{Precision + Recall}.
\end{equation}

where $Precision$ indicates precision measurement that defines the capability of a model to represent only fake statements and $Recall$ computes the aptness to refer all corresponding fake statements:

\begin{equation}
	Precision = \frac{TP}{TP+FP}.
\end{equation}

\begin{equation}
	Recall = \frac{TP}{TP+FN}.
\end{equation}
whereas ${TP}$ (True Positive) counts total number of news matched with the news in the labels. ${FP}$ (False Positive) measures the number of recognized label does not match  the annotated corpus dataset. ${FN}$ (False Negative) counts the number of news that does  not match  the predicted label news.

\item \textbf{PHEME: } Accuracy is one of the common evaluation metric to measure the performance of fake news detection on this dataset~\cite{oshikawa2018survey}. However, we also evaluate the performance based on the Fscore since our task on PHEME datasets is the binary text classification with imbalanced data. Specifically, as we perform leave-one-event-out cross-validation on the PHEME dataset, we utilize macro-averaged Fscore \cite{yang2001study} to evaluate the whole performance of mining fake news on different events.

\begin{equation}
	MacroF = \frac{1}{T} \sum_{t=1}^{T} Fscore_t.
\end{equation}
\begin{equation}
	MacroP= \frac{1}{T} \sum_{t=1}^{T} Precision_t.
\end{equation}
\begin{equation}
	MacroR= \frac{1}{T} \sum_{t=1}^{T} Recall_t.
\end{equation}
where $T$ denotes the total number of events and $Fscore_t$, $Precision_t$, $Recall_t$ are $Fscore$, $Precision$, $Recall$ values in the $t^{th}$ event. Additionally, we use macro-average accuracy on five events to examine performance. The main goal for learning from imbalanced datasets is to improve the recall without hurting the precision. However, recall and precision goals can be often conflicting, since when increasing the true positive (TP) for the minority class (True), the number of false positives (FP) can also be increased; this will reduce the precision \cite{chawla2009data}. It means that when the MacroP is increased, the MacroR might be decreased for the case of PHEME.

\begin{equation}
	MacroA= \frac{1}{T} \sum_{t=1}^{T} Accuracy_t.
\end{equation}

\end{itemize}

\subsection{Results and Discussion}

We compare the two-path deep  semi-supervised learning (TDSL) implemented based on Word CNN with the baselines on two datasets: LIAR and PHEME. In addition, we examine the effects on the performance when applying different hyper-parameters. All evaluation results are average values on five runs.

\subsubsection{LIAR} Table \ref{tab1_LIAR} presents the performance comparison  on LIAR datasets. When we focus on the baselines, Word CNN outperforms other baselines with respect to the accuracy, recall, and Fscore. However, with respect to Precision, Att RNN is better than other baselines. It means we should select specific models when we concern certain evaluation metrics. In addition, we present results generated by the proposed TDSL, where we utilize 1\% and 30\% of labeled data and the rest of unlabeled data to build the deep semi-supervised model. It is observed that the performance is strengthened  by increasing the amount of the labeled data for training TDSL. Specifically, even we use little amount of labeled data, we still obtain acceptable performance. For example, we use 1\% labeled training data to construct Word CNN based TDSL, compared with the Word CNN, its accuracy and Fscore are only reduced about 6\% and 3\%, respectively. Moreover, the accuracy, recall and Fscore of TDSL (1\%) are better than those of some baselines such as Char CNN, RCNN, Word RNN, and Att RNN. It means that the deep semi-supervised learning built based on the proposed framework is able to detect fake news even with little labeled data.

In table \ref{tab2_LIAR}, we show how the ratio of labeled data affect the detection performance generated with TDSL. When we increase the ratio of labeled data step by step, the performances such as accuracy, recall, and Fscore  are improved significantly while only precision is relatively stable. Moreover, when we use the 30\% labeled data, the performance is similar to the-state-of-the-art obtained with Word CNN. Specifically, the precision is decreased slightly when increasing the ratio of labeled data. The supervised loss, cross entropy, is defined in terms of the prediction accuracy. Therefore,  we are able to gain higher accuracy when adding more labeled data into the training procedure, but there is no guarantee to increase precision.

\begin{figure*} [ht]
	\includegraphics[scale=0.5]{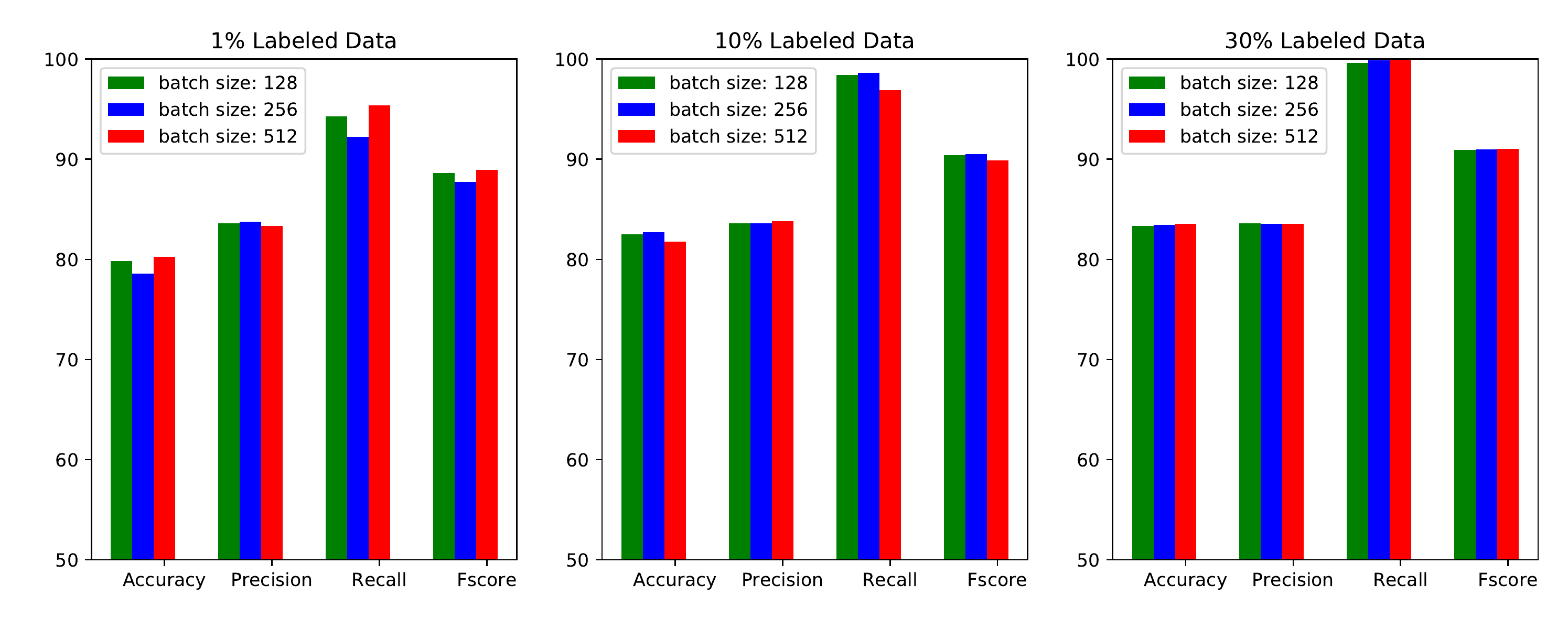}
	\centering
	\caption{Different performances generated with three batch sizes, 128, 256, and 512 on three ratios of labeled data, namely 1\%, 10\%, and 30\%. x-axis is for different evaluation metrics while y-axis is for performance. Different color bars illustrate different batch sizes, where green bars are for batch size 128, blue bars are for batch size 256, and red bars are for batch size 512.}
	\label{Fig_liar_batch_size}
\end{figure*}

\begin{figure*} [ht]
	\includegraphics[scale=0.5]{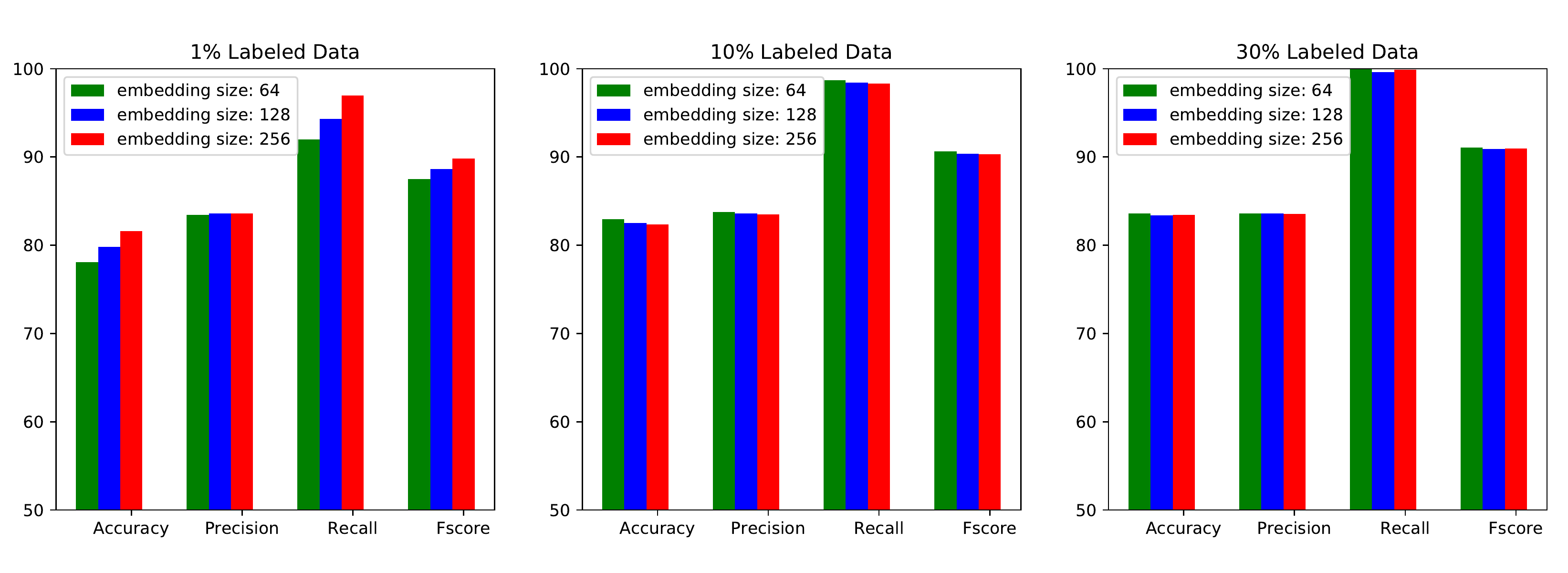}
	\centering
	\caption{Different performances generated with three embedding sizes, 64, 128, and 256 on three ratios of labeled data, namely 1\%, 10\%, and 30\%. x-axis is for different evaluation metrics while y-axis is for performance. Different color bars show different batch sizes, where green bars are for embedding size 64, blue bars are for embedding size 128, and red bars are for embedding size 256.}
	\label{Fig_liar_embedding_size}
\end{figure*}

\begin{figure*} [ht]
	\includegraphics[scale=0.5]{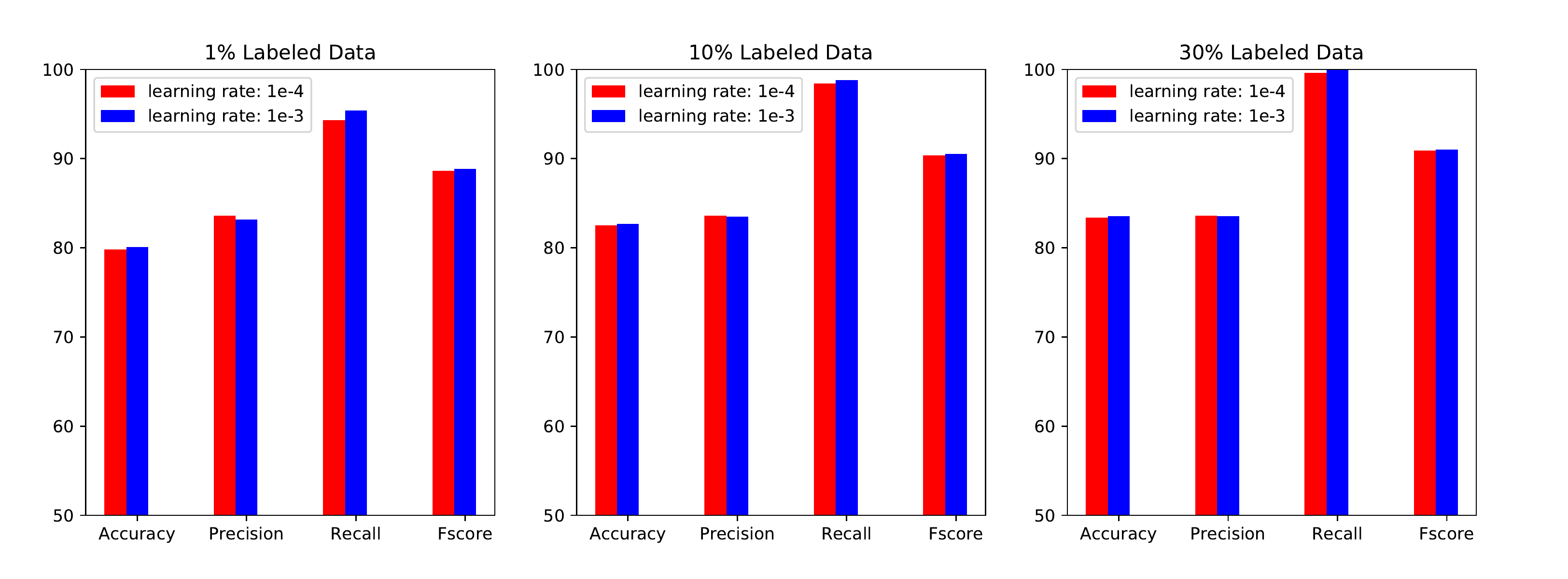}
	\centering
	\caption{Different performances generated with three learning rate, 1e-3 and 1e-4 on three ratios of labeled data, namely 1\%, 10\%, and 30\%. x-axis is for different evaluation metrics while y-axis is for performance. Different color bars indicate different batch sizes, where blue bars are for learning rate 1e-3, and red bars are for learning rate 1e-4.}
	\label{Fig_liar_learning_rate}
\end{figure*}

Additionally, we examine the performance effects from different hyper-parameters in Figure \ref{Fig_liar_batch_size}, \ref{Fig_liar_embedding_size}, and \ref{Fig_liar_learning_rate}. In Figure \ref{Fig_liar_batch_size}, we focus on the effects from different batch sizes. When we train the model with fewer labeled data, the different batch sizes affect the performance more significantly. For instance, the performance hardly change when there are 30\% labeled data while the performance varies when there are only 1\% labeled data. 
It is because more information on the ground truth will be embedded in the training samples when batch size is large, which results in robust prediction.

For Figure~\ref{Fig_liar_embedding_size} and \ref{Fig_liar_learning_rate}, we examine the performance effects on different embedding sizes and learning rates. In Figure \ref{Fig_liar_embedding_size}, there are similar observations to those of the case of batch size. For example, for the case of more labeled data for training, different embedding sizes doesn't affect the performance significantly. On the contrary, for the fewer labeled data, choosing embedding size should be carefully because different embedding sizes will lead to different performances. In addition, larger embedding size enhances the performance. In Figure \ref{Fig_liar_learning_rate}, we observe that there is no significant difference when using different learning rates in the case of 10\% labeled data and 30\% labeled data while only small performance difference can be observed in the case of 1\% labeled data.


\begin{table}
	\caption{\label{tab1_LIAR} Comparing performance between baselines and proposed model (TDSL) on LIAR Datasets. The baselines, namely, Word CNN, Char CNN, VD CNN, RCNN, WORD RNN, and Att RNN, are built with the training data that is fully labeled. On the contrary, we only apply 1\% and 30\% labeled training data and rest of unlabeled training data to accomplish learning of the proposed model. } 
        \begin{center}
                \begin{tabular}{|l|rrrr|}
                    \hline \textbf{Model} & \textbf{Accuracy} & \textbf{Precision} & \textbf{Recall} & \textbf{Fscore}\\ \hline
                    	   Word CNN\cite{kim2014convolutional}	&  \textbf{85.01\%}	& 83.57\%		& 99.94\%		&  \textbf{91.02\%}  \\
                            Char CNN)\cite{zhang2015character}	&  77.93\%  	& 84.09\%		& 91.41\%		&  87.59\%\\
                            VD CNN\cite{conneau2016very}		&  83.77\%	& 83.56\%		& 99.88\%		&  91.00\%\\
                            RCNN\cite{lai2015recurrent}			&  79.30\%	& 84.19\%		& 89.66\%		&  86.81\%\\
                            Word RNN						&  72.44\%	& 84.19\%		& 81.52\%		&  82.79\%\\ 
                            Att RNN\cite{zhou2016attention}		&  75.90\%	& 84.26\%		& 86.79\%		&  85.52\%\\  
                     \hline
                            \textbf{TDSL (1\%)}		&  \textbf{79.81\%}		&  \textbf{83.62\%}		&  \textbf{94.30\%}		&  \textbf{88.62\%}\\
                            \textbf{TDSL (30\%)}		&  \textbf{83.36\%}		&  \textbf{83.59\%}		&  \textbf{99.64\%}		&  \textbf{90.91}\%\\  
                     \hline
                \end{tabular}
       \end{center}
         \label{tab_detail}
\end{table}

\begin{table}
	\caption{\label{tab2_LIAR} Comparing performances generated by proposed model (TDSL) learning on different ratios of labeled training data and rest of unlabeled training data.}
       
        \begin{center}
                \begin{tabular}{|c|cccc|}
                    \hline \textbf{Ratio of Labeled Data} & \textbf{Accuracy} & \textbf{Precision} & \textbf{Recall} & \textbf{Fscore}\\ \hline
                    	   1\%			&  79.81\%	& 83.62\%		& 94.30\%		&  88.62\%  \\
                            3\%			&  80.71\%  	& 83.69\%		& 95.54\%		&  89.21\%\\
                            5\%			&  80.74\%	& 83.70\%		& 95.59\%		&  89.23\%\\
                            8\%			&  82.24\%	& 83.56\%		& 98.02\%		&  90.21\%\\
                            10\%			&  82.52\%	& 83.58\%		& 98.41\%		&  90.39\%\\ 
                            30\%			&  83.36\%	& 83.59\%		& 99.84\%		&  90.91\%\\  
                     \hline
                            
                \end{tabular}
       \end{center}
         \label{tab_detail}
\end{table}

\begin{table}
	\caption{\label{tab1_PHEME} Comparing performance between baselines and proposed model (TDSL) on PHEME Datasets. The baselines, namely, Word CNN, Char CNN, VD CNN, RCNN, WORD RNN, and Att RNN, are built with the training data that is fully labeled. On the contrary, we only apply 1\% and 30\% labeled training data and rest of unlabeled training data to accomplish learning of the proposed model. }
       
        \begin{center}
                \begin{tabular}{|l|cccc|}
                    \hline \textbf{Model} & \textbf{MacroA} & \textbf{MacroP} & \textbf{MacroR} & \textbf{MacroF}\\ \hline
                    	   Word CNN\cite{kim2014convolutional}	&  61.75\%	& \textbf{50.82\%}		& 17.60\%		&  24.03\%  \\
                            Char CNN)\cite{zhang2015character}	&  63.68\%  	& 50.66\%		& 19.91\%		&  26.73\%\\
                            VD CNN	\cite{conneau2016very}		&  \textbf{65.42\%}	& 49.21\%		& \textbf{30.04\%}		&  \textbf{28.50}\%\\
                            RCNN\cite{lai2015recurrent}			&  60.62\%	& 45.86\%		& 16.40\%		&  22.24\%\\
                            Word RNN						&  59.70\%	& 45.57\%		& 22.89\%		&  28.22\%\\ 
                            Att RNN\cite{zhou2016attention}		&  60.32\%	& 45.58\%		& 25.49\%		&  31.15\%\\  
                     \hline
                            \textbf{TDSL (1\%)}		&  \textbf{56.19\%}		&  \textbf{38.83\%}		&  \textbf{18.73\%}		&  \textbf{21.13\%}\\
                            \textbf{TDSL (30\%)}		&  \textbf{60.64\%}		&  \textbf{41.14\%}		&  \textbf{4.77\%}		&  \textbf{6.75}\%\\  
                     \hline
                \end{tabular}
       \end{center}
         \label{tab1_PHEME}
\end{table}

\begin{table}
	\caption{\label{tab2_PHEME} Comparing performances generated by proposed model (TDSL) learning on different ratios of labeled training data from PHEME Datasets.}
       
        \begin{center}
                \begin{tabular}{|c|cccc|}
                    \hline \textbf{Labeled Ratio} & \textbf{MacroA} & \textbf{MacroP} & \textbf{MacroR} & \textbf{MacroF}\\ \hline
                    	   1\%			&  56.19\%	& 38.83\%		& 18.73\%		&  21.13\%  \\
                            3\%			&  58.58\%  	& 39.38\%		& 13.12\%		&  17.83\%\\
                            5\%			&  58.40\%	& 39.18\%		& 12.58\%		&  16.31\%\\
                            8\%			&  59.74\%	& 40.48\%		& 8.11\%		&  11.18\%\\
                            10\%			&  59.84\%	& 40.38\%		& 7.08\%		&  10.49\%\\ 
                            30\%			&  60.64\%	& 41.14\%		& 4.77\%		&  6.75\%\\  
                     \hline
                            
                \end{tabular}
       \end{center}
         \label{tab2_PHEME}
\end{table}

\begin{table}
	\caption{\label{tab_PHEME_batch_size} Comparing performance with different batch sizes on PHEME Datasets. We choose three cases of ratios of labeled training data, namely, 1\%, 10\%, and 30\%.}
       
        \begin{center}
                \begin{tabular}{|c|cccc|}
                    \hline  &  & 1\% Labeled Data &  & \\ 
                    \hline \textbf{Batch size} & \textbf{MacroA} & \textbf{MacroP} & \textbf{MacroR} & \textbf{MacroF}\\ \hline
                    	   128			&  56.19\%	& 38.83\%		& 18.73\%		&  21.13\%  \\
                            256			&  57.78\%  	& 39.72\%		& 18.50\%		&  23.52\%\\
                            512			&  57.78\%	& 39.07\%		& 15.73\%		&  20.43\%\\
                     \hline
                    \hline  &  & 10\% Labeled Data &  & \\
                    \hline \textbf{Batch size} & \textbf{MacroA} & \textbf{MacroP} & \textbf{MacroR} & \textbf{MacroF}\\ \hline
                    	   128			&  59.84\%	& 40.38\%		& 7.08\%		&  10.49\%\ \\
                            256			&  60.08\%  	& 40.46\%		& 8.55\%		&  12.49\%\\
                            512			&  59.02\%	& 40.81\%		& 12.96\%		&  17.18\%\\
                      \hline
                     \hline  &  & 30\% Labeled Data &  & \\ 
                      \hline \textbf{Batch size} & \textbf{MacroA} & \textbf{MacroP} & \textbf{MacroR} & \textbf{MacroF}\\ \hline
                    	   128			&  60.64\%	& 41.14\%		& 4.77\%		&  6.75\%  \\
                            256			&  60.59\%  	& 41.50\%		& 7.09\%		&  10.77\%\\
                            512			&  59.94\%	& 42.77\%		& 8.51\%		&  12.27\%\\
                   \hline
                \end{tabular}
       \end{center}
         \label{tab_PHEME_batch_size}
\end{table}

\begin{table}
	\caption{\label{tab_PHEME_embedding_size} Comparing performance with different embedding sizes on PHEME Datasets. We choose three cases of ratios of labeled training data, namely, 1\%, 10\%, and 30\%.}
       
        \begin{center}
                \begin{tabular}{|c|cccc|}
                    \hline  &  & 1\% Labeled Data &  & \\ 
                    \hline \textbf{Embedding size} & \textbf{MacroA} & \textbf{MacroP} & \textbf{MacroR} & \textbf{MacroF}\\ \hline
                    	   64				&  57.38\%	& 39.20\%		& 16.57\%		&  20.78\%  \\
                            128			&  56.19\%	& 38.83\%		& 18.73\%		&  21.13\%\\
                            256			&  57.13\%	& 38.86\%		& 18.07\%		&  22.23\%\\
                     \hline
                    \hline  &  & 10\% Labeled Data &  & \\
                    \hline \textbf{Embedding size} & \textbf{MacroA} & \textbf{MacroP} & \textbf{MacroR} & \textbf{MacroF}\\ \hline
                    	   64				&  59.73\%	& 40.44\%		&  7.91\%		&  11.06\%  \\
                            128			&  59.84\%	& 40.38\%		& 7.08\%		&  10.49\%\\
                            256			&  58.89\%	& 40.27\%		& 11.67\%		&  15.22\%\\
                      \hline
                     \hline  &  & 30\% Labeled Data &  & \\ 
                      \hline \textbf{Embedding size} & \textbf{MacroA} & \textbf{MacroP} & \textbf{MacroR} & \textbf{MacroF}\\ \hline
                    	   64				&  60.79\%	& 42.32\%		& 4.37\%		&  6.99\%  \\
                            128			&  60.64\%	& 41.14\%		& 4.77\%		&  6.75\%\\
                            256			&  60.63\%	& 42.54\%		& 4.63\%		&  6.66\%\\
                   \hline
                \end{tabular}
       \end{center}
         \label{tab_PHEME_embedding_size}
\end{table}

\subsubsection{PHEME} Compared to  LIAR dataset, PHEME datasets will introduce new challenges such as imbalanced class distribution and word distribution differences among these events. Therefore, it will lead to different observations, compared to the case of LIAR. Table \ref{tab1_PHEME} indicates the  performance comparison on PHEME datasets. When we examine the baselines, VD CNN outperforms other baselines with respect to the MacroA, MacroR, and MacroF. However, considering MacroP, Word CNN is better than other baselines. In addition, we observe that the performance (MacroA) is enhanced when increasing the ratio of the labeled data for training TDSL. Moreover, even we use little amount of labeled data, we still obtain acceptable performance. For example, we use 1\% labeled training data to construct Word CNN based TDSL, compared with the VD CNN, its MacroA and MacroF are just reduced about 9\% and 7\%, respectively. However, the MacroF is decreased when MacroA is increased when adding more labeled data for training. There are two reasons for this observation. One is that the learning of TDSL aims to optimize the accuracy, but not the Fscore. The other is that the data distribution of training data is different from that of testing data since we utilize the leave-one-out policy to complete the validation, which breaks the assumption that the training data should share the same distribution to the testing data. The more labeled data is, the more serious the difference on the distribution is. 

Similar to the case of LIAR, in table \ref{tab2_PHEME}, we illustrate how the ratio of labeled data affects the detection performance. When we increase the ratio of labeled data step by step, the MacroA is improved as well,  but the MacroF is reduced significantly. Specifically, MacroR and MacroF are decreased significantly when increasing the ratio of labeled data. The supervised loss, cross entropy, is defined in terms of the prediction accuracy. Therefore, there is no guarantee to increase MacroR and MacroF when adding more labeled data into the training procedure.

In table \ref{tab_PHEME_batch_size} and \ref{tab_PHEME_embedding_size}, we examine the performance differences when choosing different batch sizes and embedding sizes, respectively. We observe the similar trends regarding the MacroA and MacroP, where there is no big difference on MacroA and MacroP when choosing different batch sizes and embedding sizes for building the proposed model. However, MacroR  is changed more significantly by different batch sizes when comparing to the case of embedding sizes.

\begin{figure*} [ht]
	\centering
	\includegraphics[scale=0.4]{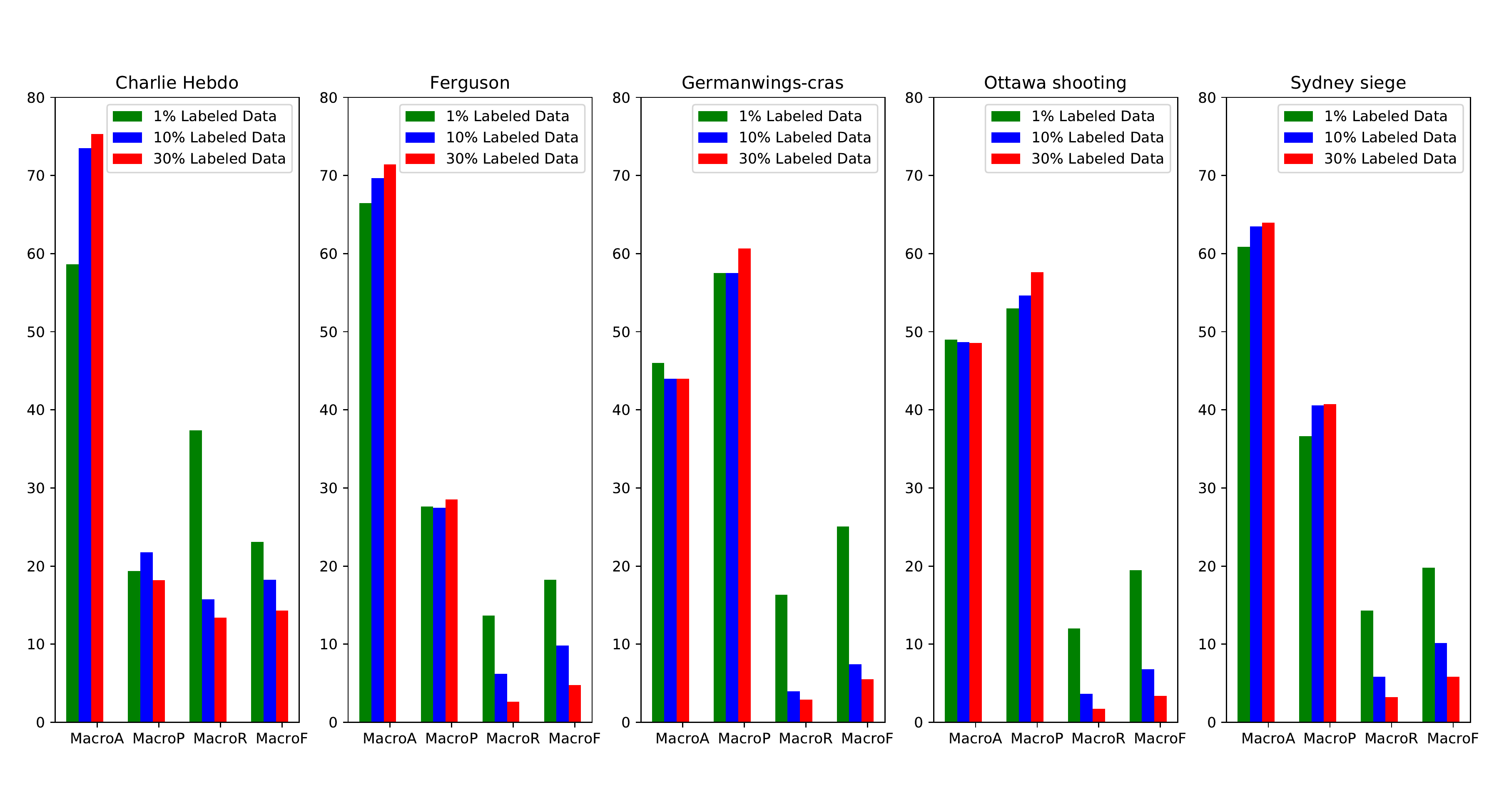}
	\caption{Comparing detailed performances generated with batch size 128 for five events. x-axis is for different evaluation metrics while y-axis is for performance. Different color bars are for different ratios of labeled data, where green bars are for 1\%, blue bars are for 10\%, and red bars are for 30\%}
	\label{Fig_pheme_batch_size_128}
\end{figure*}

\begin{figure*} [ht]
	\centering
	\includegraphics[scale=0.4]{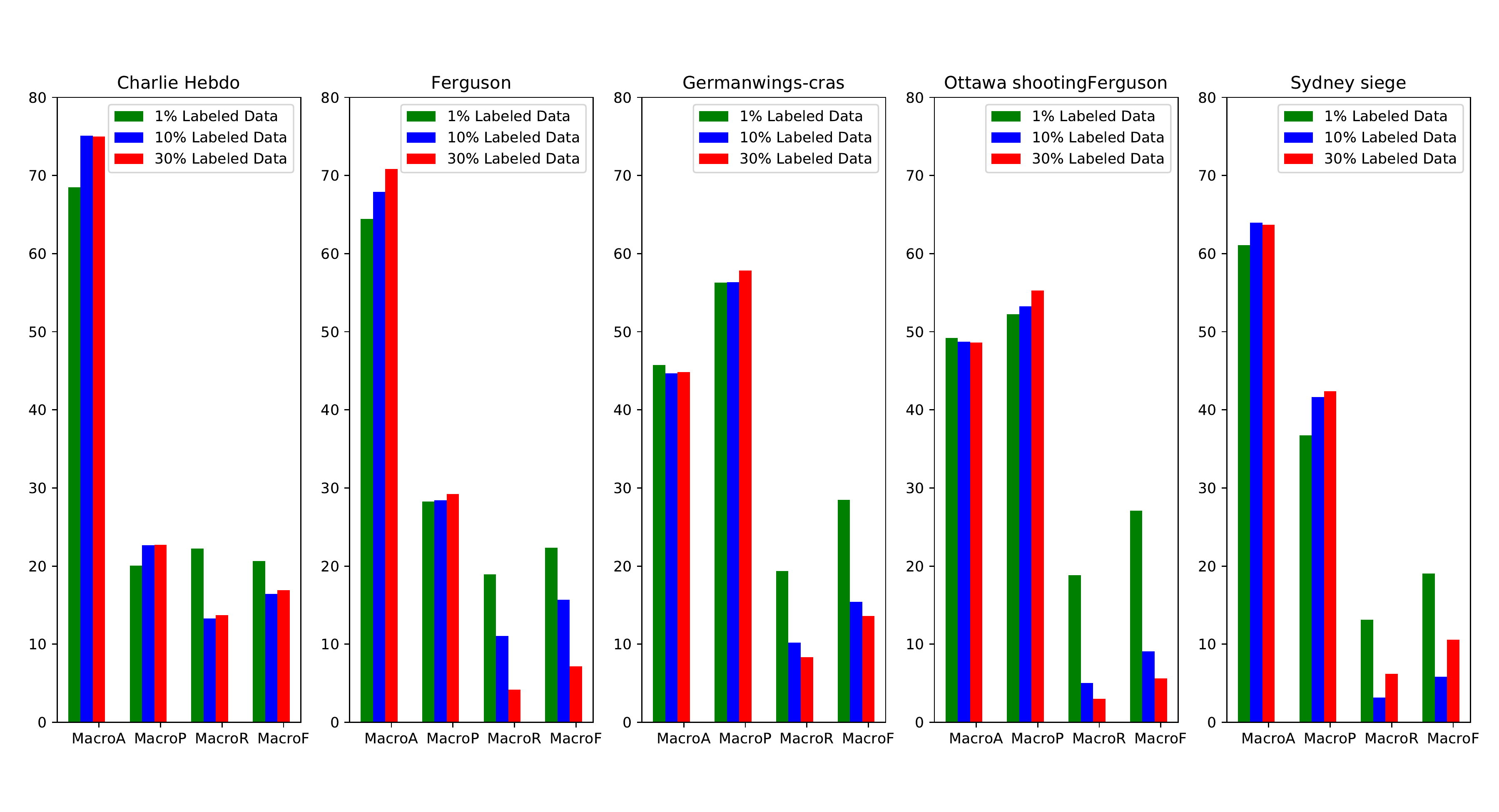}
	\caption{Comparing detailed performances generated with batch size 256 for five events. x-axis is for different evaluation metrics while y-axis is for performance. Different color bars show different ratios of labeled data, where green bars are for 1\%, blue bars are for 10\%, and red bars are for 30\%.}
	\label{Fig_pheme_batch_size_256}
\end{figure*}

\begin{figure*} [ht]
	\centering
	\includegraphics[scale=0.4]{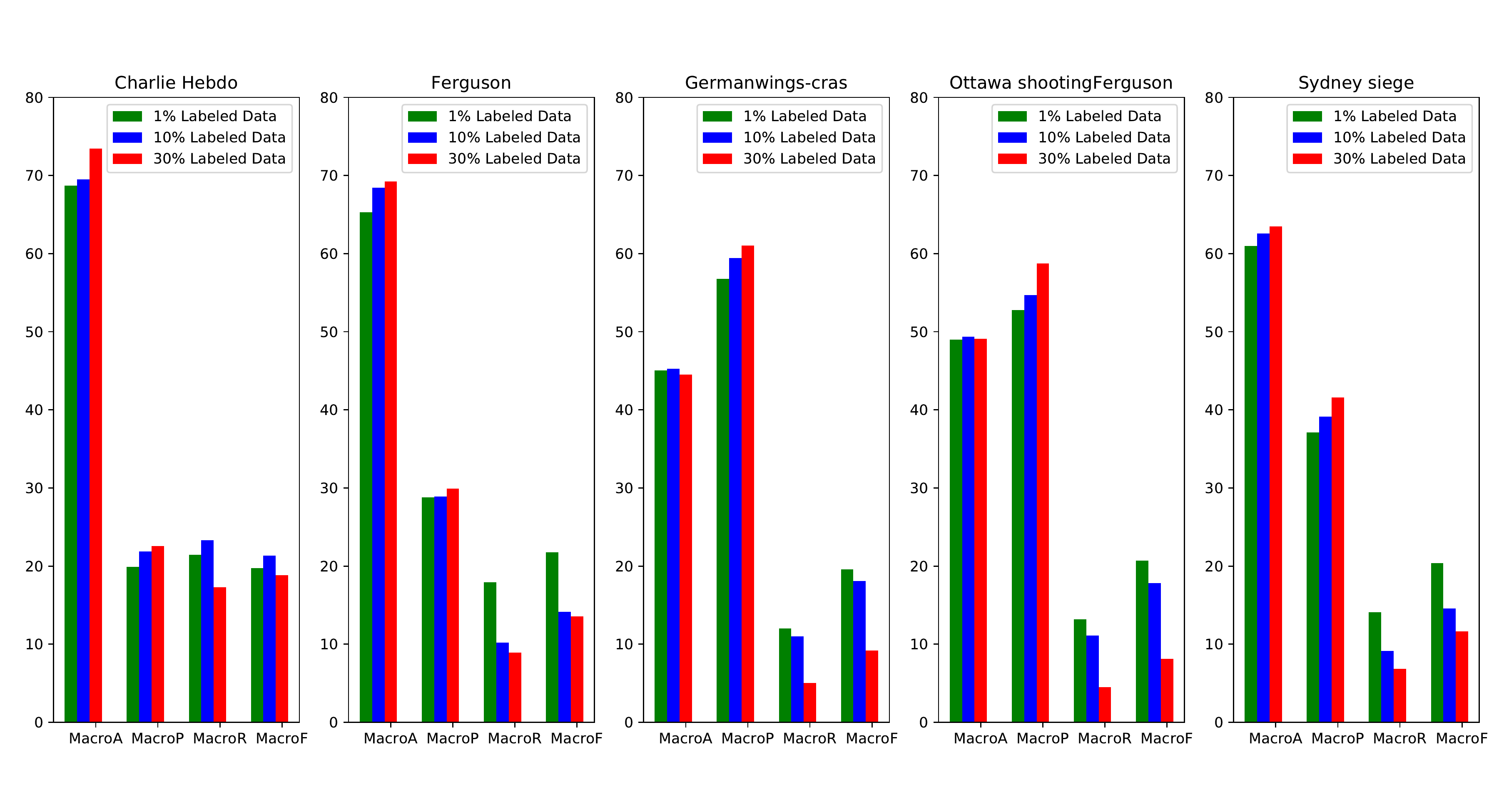}
	\caption{Comparing detailed performances generated with batch size 512 for five events. x-axis is for different evaluation metrics while y-axis is for performance. Different color bars indicate different ratios of labeled data, where green bars are for 1\%, blue bars are for 10\%, and red bars are for 30\%.}
	\label{Fig_pheme_batch_size_512}
\end{figure*}

Moreover, in the Figure \ref{Fig_pheme_batch_size_128}, \ref{Fig_pheme_batch_size_256}, and \ref{Fig_pheme_batch_size_512}, we show the detailed performances for five events when choosing different batch sizes to train the model on different ratios of labeled data. When examining the results shown in Figure  \ref{Fig_pheme_batch_size_128}, MacroA is increased for these events  Charlie Hebdo, Sydney siege, and Ferguson when increasing the amount of labeled data  whereas for the events Germanwings-cras and Ottawa shooting, the MacroA is decreased. It is because more imbalanced classes involved in the training procedure will lead to reducing the performance.  For MacroR and MacroF, the performance is reduced when adding the ratios of labeled data. In addition, the similar observations can be obtained in terms of results shown in \ref{Fig_pheme_batch_size_256}, and \ref{Fig_pheme_batch_size_512}. However, the difference is that larger batch sizes can reduce the performance affections that are from batch sizes. 

\begin{figure*} [ht]
	\centering
	\includegraphics[scale=0.4]{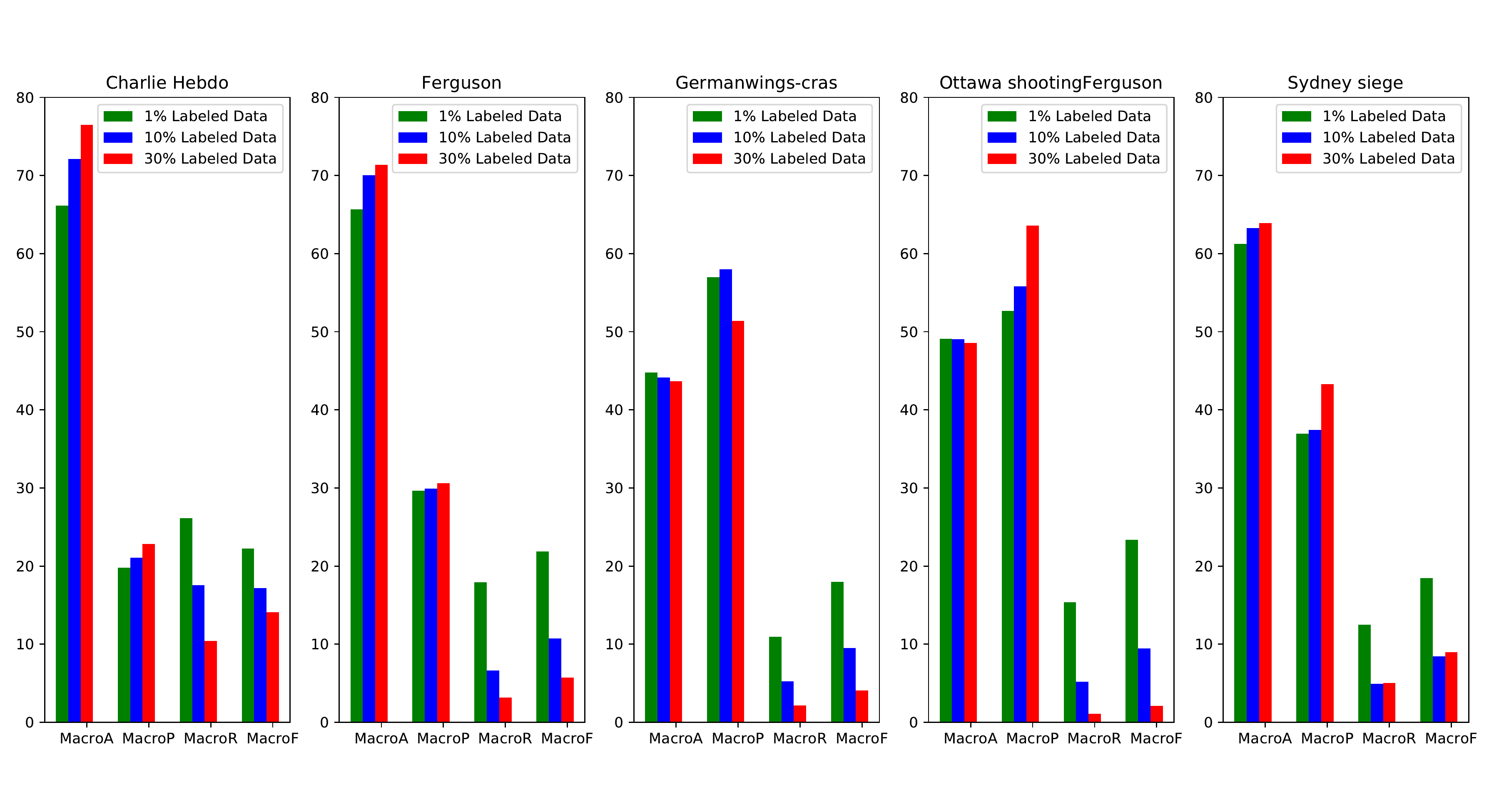}
	\caption{Comparing performance for the case of embedding size 64. x-axis is for different evaluation metrics while y-axis is for performance. Different color bars present different ratios of labeled data, where green bars are for 1\%, blue bars are for 10\%, and red bars are for 30\%.}
	\label{Fig_pheme_embedding_size_64}
\end{figure*}

\begin{figure*} [ht]
	\centering
	\includegraphics[scale=0.4]{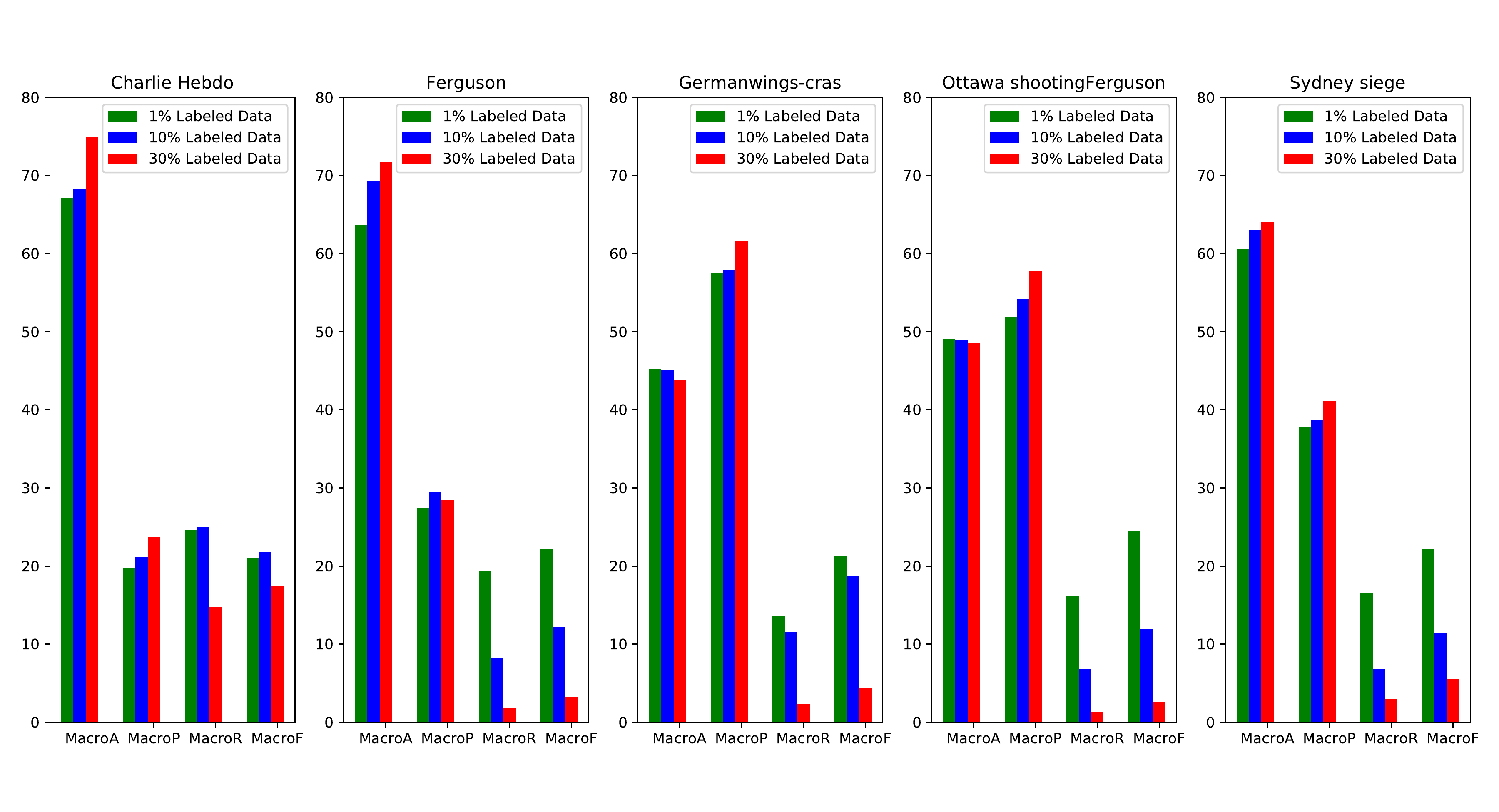}
	\caption{Comparing performance for the case of embedding size 256. x-axis is for different evaluation metrics while y-axis is for performance. Different color bars illustrate different ratios of labeled data, where green bars are for 1\%, blue bars are for 10\%, and red bars are for 30\%.}
	\label{Fig_pheme_embedding_size_256}
\end{figure*}

Finally, we examine the performance differences when choosing two different embedding sizes, namely 64 and 256, where the results are shown in Figure \ref{Fig_pheme_embedding_size_64} and \ref{Fig_pheme_embedding_size_256}, respectively. MacroA and MacroP are increased for the events  Charlie Hebdo, Sydney siege, and Ferguson when increasing the amount of labeled data for training models whereas for the events Germanwings-cras and Ottawa shooting, the MacroA is decreased. It is caused by the same reason of the case of batch size.  

In summary, in terms of observations that are from the aforementioned  results, increasing the amount of labeled data for training TDSL will enhance performance when training data has the similar distribution to the testing data, for instance, the case of LIAR. In addition, for both of benchmarks, we can obtain acceptable performance even using extremely limited labeled data for training. However, we should pay more attention to choosing the ratio of labeled when processing imbalance classification task, for example, the case of PHEME. Meantime, we should delicately choose the hyper-parameters if we plan to obtain reasonable performance.

\section{Conclusion and Future Work}
\label{sec7}

In this paper, a novel framework of deep semi-supervised learning is proposed for fake news detection. 
Because of the fast propagation of fake news, timely detection is critical to mitigate their effects. However, usually very few data samples can be labeled in a short  time, which in turn makes the supervised learning models infeasible. Hence, 
a deep semi-supervised learning model is implemented based on the proposed framework. The two paths in the model generate supervised loss (cross-entropy) and unsupervised loss (Mean Squared Error), respectively. Then training is performed by jointly optimizing these two losses. 
Experimental results indicate that the implemented model could detect fake news from these two benchmarks, LIAR and PHEME effectively using very limited labeled data and a large amount of unlabeled data. Furthermore, given the data distribution differences between training and testing datasets for the case of PHEME, and using the leave-one-event-out cross-validation, increasing the percentage of labeled data to train the semi-supervised model does not automatically imply performance improvement. In the future, we plan to examine the proposed framework on other NLP tasks such as sentiment analysis and dependency analysis.

\section*{Acknowledgment}
\label{acknowledgement}
This research work is supported in part by the U.S. Office of the Under Secretary of Defense for Research and Engineering (OUSD(R\&E)) under agreement number FA8750-15-2-0119. The U.S. Government is authorized to reproduce and distribute reprints for governmental purposes notwithstanding any copyright notation thereon. The views and conclusions contained herein are those of the authors and should not be interpreted as necessarily representing the official policies or endorsements, either expressed or implied, of the Office of the Under Secretary of Defense for Research and Engineering (OUSD(R\&E)) or the U.S. Government.

\ifCLASSOPTIONcaptionsoff
  \newpage
\fi


\bibliographystyle{IEEEtran}
\bibliography{References}
\end{document}